\documentclass[letterpaper]{article} 
\usepackage{aaai25}  
\usepackage{times}  
\usepackage{helvet}  
\usepackage{courier}  
\usepackage[hyphens]{url}  
\usepackage{graphicx} 
\usepackage{natbib}  
\usepackage{caption} 
\usepackage{algorithm}
\usepackage{algorithmic}

\usepackage{newfloat}
\usepackage{listings}
\DeclareCaptionStyle{ruled}{labelfont=normalfont,labelsep=colon,strut=off} 
\floatstyle{ruled}
\newfloat{listing}{tb}{lst}{}
\floatname{listing}{Listing}

\nocopyright 

\newcommand{\model}{LEAP}

\newcommand{\tasktwo}{MEF}
\usepackage{booktabs}
\usepackage{amssymb}
\usepackage{tabularray}
\usepackage{xcolor}
\usepackage{textcomp}
\usepackage{amsmath}

\title{Large Language Models as Interpolated and Extrapolated Event Predictors}

\author{
    Libo Zhang and Yue Ning\\
}
\affiliations{
    Department of Computer Science\\
    Stevens Institute of Technology \\
    lzhang6@stevens.edu, yue.ning@stevens.edu
}

\begin{document}

\maketitle

\begin{abstract}
Salient facts of sociopolitical events are distilled into quadruples following a format of \textit{subject, relation, object} and \textit{timestamp}. Machine learning methods, such as graph neural networks (GNNs) and recurrent neural networks (RNNs), have been built to make predictions and infer relations on the quadruple-based knowledge graphs (KGs). 

In many applications, quadruples are extended to quintuples with auxiliary attributes such as text summaries that describe the quadruple events.
In this paper, we comprehensively investigate how large language models (LLMs) streamline the design of event prediction frameworks using quadruple-based or quintuple-based data while maintaining competitive accuracy. We propose \model, a unified framework that leverages large language models as event predictors. Specifically, we develop multiple prompt templates to frame the object prediction (OP) task as a standard question-answering (QA) task, suitable for instruction fine-tuning with an encoder-decoder LLM. For multi-event forecasting (MEF) task, we design a simple yet effective prompt template for each event quintuple. This novel approach removes the need for GNNs and RNNs, instead utilizing an encoder-only LLM to generate fixed intermediate embeddings, which are processed by a customized downstream head with a self-attention mechanism to predict potential relation occurrences in the future. Extensive experiments on multiple real-world datasets using various evaluation metrics validate the effectiveness of our approach. 
\end{abstract}

\begin{links}
    \link{Code}{https://github.com/Libo1023/LEAP}
    \link{Datasets}{https://dataverse.harvard.edu/dataverse/icews}
\end{links}

\section{Introduction}

Event prediction, including tasks such as identifying key participants in an event or forecasting the occurrence of future events, has drawn significant attention in recent years across different domains~\cite{zhao2021event,onaolapo2022event,shi2024language,ye2024mirai}. The ability to successfully predict critical elements and accurately forecast human events is invaluable for proactive decision-making, risk management, and resource optimization. However, the complexity of underlying mechanisms and the inherent uncertainty of the future make event prediction a challenging endeavor~\cite{Wasserkrug2009}. These challenges necessitate advanced models that can effectively capture and interpret the dynamic patterns of events. In the past, the machine learning community has made significant progress in event reasoning and prediction by developing new methods based on recurrent neural networks (RNNs) and graph neural networks (GNNs). RNNs excel at capturing sequential information, while GNNs are adept at modeling relational knowledge within graphs~\cite{pan2024unifying, ma2023context}. These networks have become foundational in the design of state-of-the-art (SOTA) event prediction frameworks. However, text information, such as news articles that describe events, provides rich semantic indicators and contextual backgrounds that are often underutilized. With the recent advancements in large language models (LLMs), we aim to explore their potential in understanding event contexts and inferring future events based on historical patterns and textual summaries. We identify several key challenges in leveraging LLMs for event prediction, from both interpolation and extrapolation perspectives~\cite{wang2023survey, shang2024survey}. 

\textbf{The high cost of closed-source LLMs:} For interpolated object prediction (OP), we are asked to predict the missing object in a query quintuple, given the other four elements as well as historical knowledge tracing back to a certain sequence~\cite{ma2023context}. Although nowadays generating several words looks simple \cite{ouyang2022training}, good prediction accuracy cannot be easily achieved, unless introducing SOTA commercialized LLMs and leveraging complicated prompting strategies \cite{ye2024mirai}, which consumes more tokens and leads to much more costly expenses in making frequent application programming interface (API) calls for OpenAI generative LLMs \cite{openai-pricing}.

\textbf{The constrained capability of open-source LLMs:} For extrapolated multi-event forecasting (MEF), we are asked to forecast possible relation occurrences in a future time, given only historical knowledge tracing back to a certain time window without any query information \cite{deng2020dynamic}. Compared with OP, MEF is more challenging for two reasons. First, MEF follows an extrapolation setting to predict future relations, without any auxiliary knowledge such as participating subjects, while the interpolation setting provides OP with the other four query elements, and the only thing left is to fill in the missing object blank. Second, MEF is on a daily basis while OP is on the quintuple level. Since there could be hundreds of quintuples per day \cite{DVN/28075_2015}, the number of input tokens is much more demanding for LLMs involved in MEF to yield a valid downstream prediction than those in OP. However, the maximum number of input tokens for open-source LLMs is rather limited, such as 512 tokens for RoBERTa \cite{liu2019roberta} and 1024 tokens for FLAN-T5 \cite{chung2024scaling}, as longer sequences would either be truncated by the tokenizer or jeopardize the overall performance~\cite{kanakarajan2023saama}. 

To tackle these challenges, we propose \model, a unified framework that leverages \underline{\textbf{l}}arge languag\underline{\textbf{e}} models \underline{\textbf{a}}s event \underline{\textbf{p}}redictors. Considering both interpolation and extrapolation settings, we fine-tune an encoder-only LLM, RoBERTa \cite{liu2019roberta}, and an encoder-decoder LLM, FLAN-T5 \cite{chung2024scaling}, along with optimizing various customized downstream prediction heads, to achieve competitive accuracy in both OP and MEF. Our major contributions can be summarized as follows. 

\begin{enumerate}
    \item We perform OP following two directions. First, as a ranking task, we leverage a fine-tuned encoder-only LLM, RoBERTa\textsubscript{BASE} \cite{liu2019roberta}, to handle text summary in a query quintuple, and combine output embeddings back to a structural decoder through linear projection to make a prediction. Second, as a generative task following a question-answering (QA) format, where we design various prompt templates to build up questions and set correct objects as answers. We introduce an encoder-decoder LLM, FLAN-T5\textsubscript{BASE} \cite{chung2024scaling} for sequence-to-sequence generation, and leverage instruction fine-tuning to further enhance OP accuracy. 
    \item We formulate MEF as a multi-label binary classification task. We design a simple yet effective prompt template for each event, and utilize a pre-trained RoBERTa\textsubscript{LARGE} encoder ~\cite{liu2019roberta} to obtain quintuple-level embeddings. Subsequently, we customize a downstream head with self-attention \cite{vaswani2017attention} to perform weighted feature aggregation and predict possible relation occurrences in the future. 
    \item We conduct comprehensive experiments with multiple sociopolitical ICEWS datasets~\cite{DVN/28075_2015} involving different countries and various evaluation metrics, to analyze and demonstrate the validity and effectiveness of our approaches formulated in \model.
\end{enumerate}

\section{Related Work}

Event prediction has been comprehensively studied and various kinds of frameworks have been proposed to address specific challenges including, but not limited to prediction accuracy, optimization efficiency, and reasoning interpretability \cite{pan2024unifying}. Overall, these frameworks can be divided into three categories as follows \cite{liao2023gentkg}. 

\textbf{Rule-based framework:} The key components for rule-based frameworks are a list of pre-defined strategies to locate, identify, and extract parts of event quadruples or quintuples, which are assumed to be helpful for downstream inference~\cite{liu2022tlogic}. Usually, these strategies are ranked in numbers, and the constraints defined in strategies with larger numbers tend to be more relaxing than those with smaller ones to ensure the inclusion of all possible scenarios \cite{pan2023temporal}. Without any trainable parameters, rule-based frameworks are computationally efficient and easily interpretable by strictly following a list of rules. However, the implementation of manually defined rules is an exhaustive mining process within the event dataset itself, where the achieved prediction accuracy is not as competitive as those state-of-the-art (SOTA) methods with numerous parameters~\cite{liao2023gentkg}. 

\textbf{In-context learning framework:} The key modules for in-context learning frameworks are generative large language models (LLMs) with frozen parameters \cite{dong2022survey}. Various prompt templates can be designed to include few-shot learning examples and detailed instructions for completing different event prediction tasks~\cite{lee2023temporal}. Although significant efforts for model training and downstream inference can be saved, there are two obvious shortcomings. First, for free-of-charge open-source LLMs, either they impose rather limited maximum input context length, such as 1024 tokens for BART~\cite{lewis2019bart}, 2048 tokens for LLaMA~\cite{touvron2023llama}, and 4096 tokens for Llama 2~\cite{touvron2023llama2}, or demand excessive memory space for local storage and inference, such as Llama 3.1\textsubscript{405B}~\cite{meta-llama31}, while the 8B and 70B versions are less powerful. Balancing the space-performance trade-off is always challenging for local implementations. Second, closed-source commercialized LLMs, GPT-4o~\cite{openai-gpt-4o} and Claude 3~\cite{anthropic-claude-3}, are powerful generators which support much longer input context, but given abundant sociopolitical events~\cite{DVN/28075_2015}, the cost for making application programming interface (API) calls cannot be ignored~\cite{openai-pricing}. 

\textbf{Embedding-based framework:} The key components for embedding-based frameworks are entity and relation embeddings updated by graph neural networks (GNNs) and recurrent neural networks (RNNs)~\cite{ma2023context}. Extending quadruples to quintuples, handling additional text can be tricky. Specifically, for object prediction (OP), SeCoGD~\cite{ma2023context} applies the latent Dirichlet allocation algorithm~\cite{blei2003latent}, which is pre-trained on collected and filtered text summaries, to separate quintuples into different context groups, and then leverages hypergraphs to collaborate intermediate embeddings among different groups. For multi-event forecasting (MEF), Glean \cite{deng2020dynamic} builds a word graph by calculating the point-wise mutual information \cite{church1990word} between words tokenized and filtered from collected text summaries in an event dataset. Both SeCoGD and Glean are important benchmark methods, and by making appropriate use of encoding and generative LLMs, we aim to outperform these complex end-to-end frameworks, especially in terms of prediction accuracy. 

\section{LEAP for Object Prediction}

\begin{figure*}
    \centering
    \includegraphics[scale=1.20]{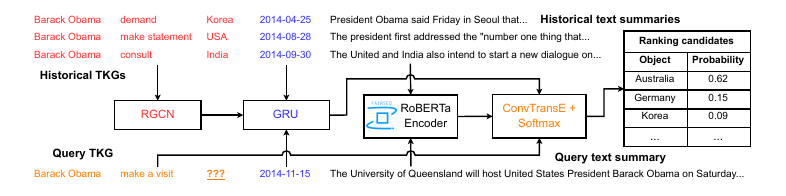}
    \caption{An overview of \model\textsubscript{OP1} for ranking object prediction. RGCN takes historical TKGs to update entity embeddings, while GRU updates relation embeddings following sequential timestamps. A fine-tuned RoBERTa\textsubscript{BASE} encodes text summaries and outputs sentence embeddings after mean pooling. These embeddings, along with the manually located query, are fed into the ConvTransE decoder to rank all object candidates.}
    \label{fig_object_prediction_ranking}
\end{figure*}

\begin{figure*}
    \centering
    \includegraphics[scale=1.20]{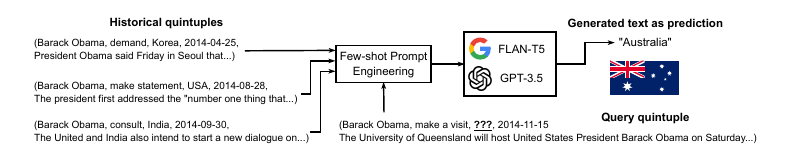}
    \caption{An overview of \model\textsubscript{OP2} for generative object prediction. Historical quintuples are concatenated as in-context learning examples during prompt engineering, and a generative LLM, either FLAN-T5\textsubscript{BASE} or GPT-3.5-Turbo-Instruct, is instructed to generate a textual prediction for the missing object entity in query.}
    \label{fig_object_prediction_generative}
\end{figure*}

Event interpolation refers to the retrieval or prediction of one or some missing facts at one specified timestamp~\cite{cai2024survey}, which corresponds to the definition of object prediction (OP) task, where we aim to predict a missing object $o_q$, given a query quintuple $(s_q, r_q, ?, t_q, x_q)$ including known subject ($s_q$), relation ($r_q$), timestamp ($t_q$), and text summary ($x_q$), as well as selected historical knowledge tracing back to a pre-defined sequence length $l$, which can be structured as a list of graphs $\mathbf{G}_{\leq t_q}=\{G_{t_q-l}, G_{t_q-l-1}, ..., G_{t_q-1}, G_{t_q}\}$, or concatenated into textual corpus formatted as multiple strings $\mathbf{X}_{\leq t_q}$~\cite{ma2023context, liao2023gentkg}. 

\subsection{\model\textsubscript{OP1}: Object Prediction as A Ranking Task}

Figure~\ref{fig_object_prediction_ranking} shows an overview of \model\textsubscript{OP1}, which can be divided into three steps starting from embedding update, then to embedding aggregation, and eventually towards probability ranking to complete the prediction. 

\textbf{1) Spatial-temporal modeling} Inspired by SeCoGD~\cite{ma2023context}, we leverage relational graph convolution network (R-GCN)~\cite{schlichtkrull2018modeling} to handle the structural knowledge for every sub-graph in $G_{\leq t_q}$ and update the embeddings for all entities $V$. We also apply gated recurrent unit (GRU)~\cite{cho2014learning} to track the historical sequence starting from $(t_q-l_1)$ and keep the embeddings of all relations $R$ up-to-date until $(t_q)$.


\textbf{2) Text summary embedding collection} To begin, we collect and concatenate summaries from all quintuples to form a large textual corpus. Subsequently, we leverage part of the corpus as training data to fine-tune an encoder-only LLM, RoBERTa\textsubscript{BASE}, with the standard masked language modeling loss \cite{liu2019roberta}, to enrich the LLM encoder with domain-specific knowledge. We apply the fine-tuned encoder towards each historical and query text summary to yield token-level embeddings. Then, we utilize mean pooling to obtain an embedding vector for each text summary. 

\textbf{3) Query object probability ranking} We leverage ConvTransE~\cite{shang2019end} as the structural decoder, which is responsible for combining the entity and relation embeddings $\mathbf{E}_{V,R}$ from R-GCN and GRU with text summary embeddings $\mathbf{E}_{text}$ from fine-tuned RoBERTa\textsubscript{BASE}, and eventually ranking all candidate objects after $\text{Softmax}(\cdot)$ activation to complete the missing query. Compared with SeCoGD \cite{ma2023context}, where the context IDs are acquired from pre-trained clustering algorithms \cite{blei2003latent}, we believe that the semantic information encoded in $\Tilde{\mathbf{E}}_{text}$ can significantly enrich event prediction optimization and improve the accuracy. Therefore, we get the probabilities of all entity candidates as:
\begin{equation}
    \mathbf{P}_q = \text{Softmax}[\text{ConvTransE}(\mathbf{E}_{V,R},\mathbf{\mathbf{E}}_{text})]\vert_{q}
\end{equation}
where $\mathbf{P}_q\in \mathbb{R}^{b\times \lvert V\rvert}$, $b$ is the query batch size, and we have $\sum_{j=1}^{\lvert V\rvert}p_{q_i}^{j}=1$, $i=1, 2, ..., b$. The candidate entity $v_j$ having the highest probability among $V$ will be read out as the predicted object. Overall, since the LLM encoder has been separately fine-tuned, the trainable modules in \model\textsubscript{OP1} contain R-GCN, GRU, and ConvTransE, which are jointly optimized with cross-entropy loss comparing predicted objects with true objects through back-propagation. 

\subsection{\model\textsubscript{OP2}: Object Prediction as A Generative Task}

\begin{table*}[t]
\centering
\small
\caption{Multiple prompt templates for generative object prediction.}
\label{table_few-shot_prompt}
\begin{tabular}{lp{14cm}}
\toprule
\textbf{Name} & \textbf{Prompt Template} \\
\midrule
Few-shot & I ask you to perform an object prediction task after I provide you with five examples. Each example is a knowledge quintuple containing two entities, a relation, a timestamp, and a brief text summary.  Each knowledge quintuple is strictly formatted as (subject entity, relation, object entity, timestamp, text summary). For the object prediction task, you should predict the missing object entity based on the other four available elements. 
 Now I give you five examples. \\
         & \#\# Example 1 \\
         & (\textcolor{blue}{$\langle$\textsc{subject 1}$\rangle$}, \textcolor{blue}{$\langle$\textsc{relation 1}$\rangle$}, $\langle$\textsc{MISSING OBJECT ENTITY}$\rangle$, \textcolor{blue}{$\langle$\textsc{timestamp 1}$\rangle$}, \textcolor{blue}{$\langle$\textsc{text summary 1}$\rangle$}) \textcolor{gray}{\textbackslash n} \\
         & The $\langle$\textsc{MISSING OBJECT ENTITY}$\rangle$ is: \textcolor{blue}{$\langle$\textsc{object 1}$\rangle$} \textcolor{gray}{\textbackslash n}\\
         & \vdots \#\# Example 5 \\
         & (\textcolor{blue}{$\langle$\textsc{subject 5}$\rangle$}, \textcolor{blue}{$\langle$\textsc{relation 5}$\rangle$}, $\langle$\textsc{MISSING OBJECT ENTITY}$\rangle$, \textcolor{blue}{$\langle$\textsc{timestamp 5}$\rangle$}, \textcolor{blue}{$\langle$\textsc{text summary 5}$\rangle$}) \textcolor{gray}{\textbackslash n} \\
         & The $\langle$\textsc{MISSING OBJECT ENTITY}$\rangle$ is: \textcolor{blue}{$\langle$\textsc{object 5}$\rangle$} \textcolor{gray}{\textbackslash n}\\
         & Now I give you a query: \\
         & (\textcolor{blue}{$\langle$\textsc{subject 6}$\rangle$}, \textcolor{blue}{$\langle$\textsc{relation 6}$\rangle$}, $\langle$\textsc{MISSING OBJECT ENTITY}$\rangle$, \textcolor{blue}{$\langle$\textsc{timestamp 6}$\rangle$}, \textcolor{blue}{$\langle$\textsc{text summary 6}$\rangle$}) \textcolor{gray}{\textbackslash n} \\
         & Please predict the missing object entity. You are allowed to predict new object entity which you have \\
         & never seen in examples. The correct object entity is: \\
\midrule
Zero-shot & Remove all five in-context learning examples in the few-shot prompt. \\
\midrule
No-text & Remove all text summaries of five in-context learning examples and the query in the few-shot prompt. \\
\bottomrule
\end{tabular}
\end{table*}

Having carefully studied a structural ranking approach, our next attempt is to evaluate if LLMs are able to predict objects directly as a generative predictor. 
In other words, we stack historical knowledge, queries, and instructions into a textual prompt context $C_{in}$, and feed $C_{in}$ into a generative LLM, such as FLAN-T5 \cite{chung2024scaling} or GPT-4o~\cite{openai-gpt-4o}. This process is denoted by
\begin{equation}
    C_{out} = \text{Generative-LLM}(C_{in})
\end{equation}
where $C_{out}$ indicates the generated context from an LLM. To make sure that the LLM generates missing objects precisely and correctly ($C_{out}=o_q$), we frame OP as a generative task. We introduce ROUGE scores \cite{lin2004rouge} to directly compare the true object and the predicted object as two strings as shown in Figure~\ref{fig_object_prediction_generative}, instead of ranking all candidate objects. 
To transfer from ranking into generation, we directly read out the first candidate from \model\textsubscript{OP1} which has the highest output probability as the predicted object. Then we go over all test queries and  compute the ROUGE scores for evaluation. We construct the generative predictor using a light-weight LLM, FLAN-T5\textsubscript{BASE},  which has around 248 million parameters~\cite{chung2024scaling}. Specifically, since FLAN-T5 is an encoder-decoder sequence-to-sequence (Seq2Seq) model, as shown in Figure~\ref{fig_object_prediction_generative}, we follow a typical question-answering format, where the ``question'' $Q_{in}$ is a 5-shot learning prompt with one query along with some explanations and instructions, and the ``answer'' $A_{out}$ is the object entity. We get
\begin{equation}
    A_{out}=\text{Encoder-Decoder-LLM}(Q_{in})
\end{equation}
where we use a fine-tuned FLAN-T5\textsubscript{BASE} as the encoder-decoder LLM  following by a standard cross-entropy loss. The purpose of fine-tuning is to further enhance the generative QA accuracy, so that \model\textsubscript{OP2} can compete with commercialized LLMs such as GPT-4o mini~\cite{openai-gpt-4o}. Based on Table~\ref{table_few-shot_prompt}, we fine-tune FLAN-T5\textsubscript{BASE} with a 5-shot learning prompt (\textit{few-shot prompt}), which means the sequence length is 5 on the quintuple level, $l_2=5$ events. To evaluate the generalizability of the encoder-decoder LLM in the experimental section, we remove the 5 learning examples to build a \textit{zero-shot prompt}. To further evaluate the effectiveness introduced by text, we design another 5-shot learning prompt (\textit{no-text prompt}) but remove all text summaries as an ablation study. Given the vast scale of the sociopolitical dataset~\cite{DVN/28075_2015}, the limited GPU memory available on our servers, and the high cost associated with commercial LLMs~\cite{openai-pricing}, we opted not to prioritize prompting strategies like Chain-of-Thought~\cite{wei2022chain} or ReAct~\cite{yao2022react}.

\section{LEAP for Multi-event Forecasting}

\begin{figure*}
    \centering
    \includegraphics[scale=1.16]{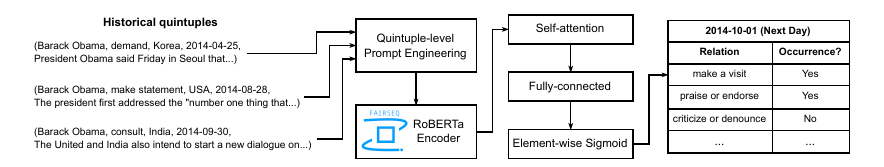}
    \caption{An overview of \model\textsubscript{MEF} for multi-event forecasting. Historical quintuples are fed into the \textit{simple prompt} template and the pre-trained RoBERTa\textsubscript{LARGE} encoder one by one, resulting in multiple quintuple-level embeddings, which are aggregated through the self-attention mechanism. Eventually, a fully connected layer with element-wise $\text{Sigmoid}(\cdot)$ activation and a threshold of 0.5 for predictions of relation occurrences in the future.}
    \label{fig_multi-event_forecasting}
\end{figure*}

\begin{table}[t]
\centering
\small
\caption{The \textit{simple prompt} template for \model\textsubscript{MEF}.}
\label{table_simple_prompt}
\begin{tabular}{ll}
\toprule
\textbf{Name} & \textbf{Prompt Template} \\
\midrule
Simple & Subject: \textcolor{blue}{$\langle$\textsc{subject}$\rangle$}; \textcolor{gray}{\textbackslash n} \\ 
       & Relation: \textcolor{blue}{$\langle$\textsc{relation}$\rangle$}; \textcolor{gray}{\textbackslash n}\\ 
       & Object: \textcolor{blue}{$\langle$\textsc{object}$\rangle$}; \textcolor{gray}{\textbackslash n}\\ 
       & Timestamp: \textcolor{blue}{$\langle$\textsc{timestamp}$\rangle$}; \textcolor{gray}{\textbackslash n}\\ 
       & Text Summary: \textcolor{blue}{$\langle$\textsc{text summary}$\rangle$}\\ 
\bottomrule
\end{tabular}
\end{table}

\begin{table}[t]
\small
\centering
\caption{ICEWS dataset statistics.}
\label{tab:data_stats}
\begin{tabular}{lccccc}
\toprule
\textbf{Dataset} & $\lvert V\rvert$ & $\lvert R\rvert$
                 & $\lvert Q\rvert_{\textbf{train}}$ 
                 & $\lvert Q\rvert_{\textbf{valid}}$ 
                 & $\lvert Q\rvert_{\textbf{test}}$ \\
\midrule
Afghanistan & 3,756 & 218 & 212,540 & 32,734 & 34,585 \\
India & 6,298 & 234 & 318,471 & 75,439 & 85,739 \\
Russia & 7,798 & 237 & 275,477 & 46,516 & 51,371 \\
\bottomrule
\end{tabular}
\end{table}

\begin{table*}[t]
\centering
\caption{Object prediction results as a ranking task.}
\label{tab:opranking}
\begin{tabular}{lccccccccccc}
\toprule 
\textbf{Method}& \multicolumn{3}{c}{\textbf{Afghanistan}} & &\multicolumn{3}{c}{\textbf{India}} & &\multicolumn{3}{c}{\textbf{Russia}} \\
\midrule
  & Hits@ 1      & Hits@ 3      & Hits@ 10  & & Hits@ 1      & Hits@ 3      & Hits@ 10    & & Hits@ 1      & Hits@ 3      & Hits@ 10 \\
\cmidrule{2-4}\cmidrule{6-8}\cmidrule{10-12}
\multicolumn{12}{c}{Historical sequence length $=$ 3 ($l_1=3$ days)} \\
\midrule
ConvTransE       & 0.1235 & 0.2704 & 0.4916 && 0.1521 & 0.2821 & 0.4600 && 0.1009 & 0.1791 & 0.3078 \\
CompGCN          & 0.1214 & 0.2633 & 0.4900 && 0.1697 & 0.3138 & 0.4940 && 0.1167 & 0.1885 & 0.3124 \\
R-GCN            & 0.1443 & 0.3126 & 0.5540 && 0.1721 & 0.3211 & 0.5009 && 0.1314 & 0.2389 & 0.3896 \\
RE-GCN           & 0.1538 & 0.3137 & 0.5408 && 0.1704 & 0.3125 & 0.4952 && 0.1332 & 0.2345 & 0.3767 \\
SeCoGD           & 0.1878 & 0.3570 & 0.5740 && 0.2064 & 0.3554 & 0.5357 && 0.1768 & 0.2909 & 0.4351 \\
\model\textsubscript{OP1} & \textbf{0.3691} & \textbf{0.5630} & \textbf{0.7317} & &
                    \textbf{0.3675} & \textbf{0.5507} & \textbf{0.7233} & &
                    \textbf{0.3751} & \textbf{0.5390} & \textbf{0.6831} \\
\midrule
\multicolumn{12}{c}{Historical sequence length $=$ 7 ($l_1=7$ days)} \\
\midrule
ConvTransE       & 0.1243 & 0.2759 & 0.4909 && 0.1544 & 0.2911 & 0.4740 && 0.0973 & 0.1804 & 0.2971 \\
CompGCN          & 0.1094 & 0.2729 & 0.5081 && 0.1655 & 0.3123 & 0.4989 && 0.1069 & 0.1851 & 0.3158 \\
R-GCN            & 0.1582 & 0.3243 & 0.5526 && 0.1762 & 0.3209 & 0.5004 && 0.1276 & 0.2462 & 0.3926 \\
RE-GCN           & 0.1551 & 0.3298 & 0.5544 && 0.1724 & 0.3175 & 0.5004 && 0.1335 & 0.2349 & 0.3781 \\
SeCoGD           & 0.1833 & 0.3652 & 0.5862 && 0.2056 & 0.3516 & 0.5352 && 0.1661 & 0.2823 & 0.4433 \\
\model\textsubscript{OP1} & \textbf{0.3861} & \textbf{0.5884} & \textbf{0.7664} & &
                    \textbf{0.3935} & \textbf{0.5831} & \textbf{0.7454} & &
                    \textbf{0.3861} & \textbf{0.5590} & \textbf{0.7077} \\
\bottomrule
\end{tabular}
\end{table*}

\begin{table*}[t]
\centering
\caption{Object prediction results as a generative task.}
\label{tab:opqa}
\begin{tabular}{lccccccccccc}
\toprule
\textbf{Method} & \multicolumn{3}{c}{\textbf{Afghanistan}} &&  \multicolumn{3}{c}{\textbf{India}} &&  \multicolumn{3}{c}{\textbf{Russia}} \\
\midrule
ROUGE$-$  & 1      & 2      & L     &  & 1      & 2      & L &      & 1      & 2      & L \\
\cmidrule{2-4}\cmidrule{6-8}\cmidrule{10-12}
\multicolumn{12}{c}{Fine-tuned RoBERTa-base as an encoding LLM, $l_1=3$ days} \\
\midrule
SeCoGD            & 0.4271 & 0.1658 & 0.4271 && 0.4812 & 0.2902 & 0.4813 && 0.3698  & 0.1414 & 0.3700 \\
\model\textsubscript{OP1} & 0.5287 & 0.2670 & 0.5286 && 0.5500 & 0.3821 & 0.5501 &&  0.4974 & 0.2629 & 0.4973 \\
\midrule
\multicolumn{12}{c}{Fine-tuned RoBERTa-base as an encoding LLM, $l_1=7$ days} \\
\midrule
SeCoGD            & 0.4295 & 0.1723 & 0.4295 &&  0.4831 & 0.2905 & 0.4833 && 0.3831 & 0.1404 & 0.3831 \\
\model\textsubscript{OP1} & 0.5675 & 0.3200 & 0.5674 &&  0.5892 & 0.4034 & 0.5893 && 0.5242 & 0.2636 & 0.5238 \\
\midrule
\multicolumn{12}{c}{Open-source LLMs: few-shot prompt for all test samples, $l_2=5$ quintuples} \\
\midrule
Mistral-7B  & 0.3659 & 0.1100 & 0.3620 && 0.3424 & 0.1318 & 0.3417 && 0.3464 & 0.1178 & 0.3452 \\
Llama3.1-8B & 0.3110 & 0.1085 & 0.3100 && 0.2799 & 0.1266 & 0.2796 && 0.3932 & 0.1575 & 0.3930 \\
\midrule
\multicolumn{12}{c}{Closed-source LLMs: few-shot prompt for the first 5000 test samples, $l_2=5$ quintuples} \\
\midrule
GPT-3.5 Turbo & 0.4097 & 0.1302 & 0.4092 && 0.3644 & 0.1601 & 0.3640 && 0.3480 & 0.1255 & 0.3485 \\
GPT-4o mini   & 0.4147 & 0.1303 & 0.4131 && 0.3839 & 0.1769 & 0.3829 && 0.3880 & 0.1726 & 0.3878 \\
\midrule
\multicolumn{12}{c}{\model\textsubscript{OP2}: fine-tuned FLAN-T5\textsubscript{BASE}, multiple prompts for all test samples, $l_2=5$ quintuples} \\
\midrule
No-text   & 0.4216 & 0.1322 & 0.4215 & & 0.4779 & 0.2517 & 0.4780 & & 0.3612 & 0.1007 & 0.3614 \\
Zero-shot & \underline{0.8601} & \underline{0.5601} & \underline{0.8600} & & 
            \underline{0.8530} & \underline{0.6885} & \underline{0.8531} & & 
            \underline{0.8318} & \underline{0.4480} & \underline{0.8317} \\
Few-shot  & \textbf{0.8638} & \textbf{0.5656} & \textbf{0.8638} & & 
            \textbf{0.8594} & \textbf{0.6962} & \textbf{0.8594} & & 
            \textbf{0.8415} & \textbf{0.4544} & \textbf{0.8414} \\
\bottomrule
\end{tabular}
\end{table*}

\begin{table*}[t]
\centering
\caption{Multi-event forecasting results, historical sequence length $=$ 7 ($l_3=7$ days).}
\label{table_MEF}
\begin{tabular}{lccccccccccc}
\toprule
\textbf{Method} & \multicolumn{3}{c}{\textbf{Afghanistan}} & & 
                   \multicolumn{3}{c}{\textbf{India}} & & 
                   \multicolumn{3}{c}{\textbf{Russia}} \\
\midrule
 & F1 & Recall & Precision & & 
         F1 & Recall & Precision & & 
         F1 & Recall & Precision \\
\cmidrule{2-4}\cmidrule{6-8}\cmidrule{10-12}
MLARAM & 0.3384 & 0.3526 & 0.3253 && 0.3368 & 0.3410 & 0.3327 && 0.2567 & 0.2671 & 0.2471 \\
BR-KNN & 0.4989 & 0.6152 & 0.4196 && 0.5036 & 0.5600 & 0.4575 && 0.4746 & 0.5664 & 0.4084 \\
ML-KNN & 0.4543 & 0.5035 & 0.4139 && 0.5233 & 0.5577 & 0.4929 && 0.5138 & 0.5862 & 0.4573 \\
DNN    & 0.5577 & 0.6814 & 0.4720 & & 0.5249 & 0.5638 & 0.4910 & & 0.5381 & 0.6261 & 0.4718 \\
Dynamic GCN  & 0.5005 & 0.5775 & 0.4416 & & 0.4180 & 0.4319 & 0.4050 & & 0.5281 & 0.6014 & 0.4707 \\
Temporal GCN & 0.6004 & 0.7693 & 0.4923 & & 0.6073 & 0.6720 & 0.5540 & & 0.5636 & 0.6766 & 0.4829 \\
RENET  & 0.6058 & 0.7775 & 0.4962 & & 0.5844 & 0.6418 & 0.5364 & & 0.5585 & 0.6566 & 0.4859 \\
Glean & \underline{0.6248} & \underline{0.8284} & \textbf{0.5015} & & 
        \underline{0.6669} & \underline{0.7731} & \underline{0.5864} & & 
        \underline{0.5892} & \underline{0.7357} & \underline{0.4914} \\
\model\textsubscript{MEF\textbackslash SA} & 0.6093 & 0.7832 & \underline{0.4986} & & 
                          0.5921 & 0.6476 & 0.5454 & & 0.5667 & 0.6838 & 0.4838 \\
\model\textsubscript{MEF}           & \textbf{0.6363} & \textbf{0.8869} & 0.4961
                       & & \textbf{0.7099} & \textbf{0.8731} & \textbf{0.5981}
                       & & \textbf{0.6280} & \textbf{0.8681} & \textbf{0.4919} \\
\bottomrule
\end{tabular}
\end{table*}

\begin{table*}[t]
\centering
\caption{Computational cost comparison, 1 epoch for training, and 1000 test samples for inference.}
\label{table_compare_cost}
\begin{tabular}{lcccccc}
\toprule
\textbf{Model} & \textbf{Task} & \textbf{Implementation} 
& \textbf{GPU memory} & \textbf{Training time} & \textbf{Inference time} & \textbf{Money spent} \\
\midrule
Mistral-7B    & OP & Local server & 15,695 MiB & / & 4 mins 23 secs & / \\
Llama-3.1-8B  & OP & Local server & 21,441 MiB & / & 2 mins 46 secs & / \\
GPT-3.5 Turbo & OP & Online API & / & / & 6 mins 12 secs & 1.00 USD \\
GPT-4o mini   & OP & Online API & / & / & 7 mins 45 secs & 0.10 USD \\
SeCoGD        & OP & Local server & 9,219 MiB & 4 mins 41 secs & $<$ 10 secs & / \\
\model\textsubscript{OP1} & OP & Local server & 13,412 MiB & 13 mins 00 sec & $<$ 10 secs & / \\
\model\textsubscript{OP2} & OP & Local server & 7,851 MiB & 9 hours 43 mins & 1 min 10 secs & / \\
Glean             & \tasktwo & Local server & 7,109 MiB & 1 min 26 secs & / & / \\
\model\textsubscript{MEF}    & \tasktwo & Local server & 10,369 MiB & 3 mins 49 secs & / & / \\
\bottomrule
\end{tabular}
\end{table*}

Event extrapolation refers to the forecasting of events of interest in a future time, given only historical data~\cite{cai2024survey}, which complies with the definition of multi-event forecasting (MEF) task. In this task, we aim to forecast possible relation occurrences $r^{i}_{t_n}\in R$, where $i=1,2,...,\lvert R\rvert$ in the next timestamp $t_{n}$, given only historical information either in a structural format $\mathbf{G}_{<t_n}$ or a textual corpus format $\mathbf{X}_{<t_n}$~\cite{deng2020dynamic}.

We start with a straightforward and simple prompt template $u_i$, as shown in Table~\ref{table_simple_prompt}, to express an event quintuples as plain language, which can be fed into an encoder-only LLM. The total number of \textit{simple prompts} is equal to the total number of quintuples $Q$ for each dataset, where $i=1,2,...,\lvert Q\rvert$~\cite{DVN/28075_2015}. Then, we leverage a pre-trained RoBERTa\textsubscript{LARGE}, along with mean pooling $\text{Mean}(\cdot)$ for token-level aggregation, to generate an embedding vector $e_i$ for each quintuple prompt $u_i$, such that
\begin{equation}
    e_i=\text{Mean}( \text{Encoder-only-LLM}(u_i))
\end{equation}
where $e_i\in \mathbb{R}^{d_{h_3}}$ and $d_{h_3}$ is the output feature dimension for RoBERTa\textsubscript{LARGE} encoder. The purpose of utilizing quintuple-level prompt engineering is to address the challenge of limited input context of open-source LLMs. Before feeding collected embeddings into downstream MEF layers, we need to convert them from the quintuple level back to the daily level with a purpose of incorporating as much history as possible for future forecasting. We write daily embeddings as $\mathbf{E}_{j}\in \mathbb{R}^{n_j\times d_{h_3}}$, where $j=1,2,...,\lvert T \rvert$, $T$ is the set of timestamps, and $n_j$ is the number of events on the $j$th day. 

Based on Figure~\ref{fig_multi-event_forecasting}, tracing back to a pre-defined sequence length $l_3$, we feed concatenated embeddings $\mathbf{E}_{t_n-l_3\leq t < t_n}$ into a single-head self-attention layer~\cite{vaswani2017attention}, and then take a mean aggregation to collapse the dimension for multiple historical days, as we get
\begin{equation}
    \Tilde{\mathbf{E}}_{t_n}=\text{Mean}[\text{self-attention}(\mathbf{E}_{t_n-l_3\leq t < t_n})]\vert_{t_n}
\end{equation}
where $\Tilde{\mathbf{E}}_{t_n}\in \mathbb{R}^{b\times d_{h_3}}$ indicates the aggregated embeddings and $b$ is the batch size for future days, $t_n=1,2,...,b$. 

Subsequently, we introduce a fully-connected layer $z(\cdot)$ to adjust feature dimension and forecast possible relations in the future $t_n$, following an element-wise $\text{Sigmoid}(\cdot)$ with a threshold 0.5 to decide occurrences. We get 
\begin{equation}
    \mathbf{P}_{t_n}=\text{Sigmoid}(z(\Tilde{\mathbf{E}}_{t_n}))
\end{equation}
where $\mathbf{P}_{t_n}\in \mathbb{R}^{b\times \lvert R\rvert}$, and $R$ is the set of all unique relations. $z(\cdot)$ contains a trainable matrix $W_z\in \mathbb{R}^{d_{h_3}\times \lvert R\rvert}$ and vectorized bias $b_z\in \mathbb{R}^{\lvert R\rvert}$. Overall, for \model\textsubscript{MEF}, the trainable modules are the self-attention layer and the fully-connected layer $z(\cdot)$, which are optimized with cross-entropy loss comparing predicted relation occurrences with true occurrences through back-propagation. 

\section{Experiments}

We conduct experiments on event datasets of three countries (Afghanistan, India, and Russia) from ICEWS which contains sociopolitical facts structured for crisis analysis, and each event is further enriched with several textual sentences sliced from authorized news articles~\cite{DVN/28075_2015}. 
Table~\ref{tab:data_stats} shows basic data statistics. The timestamps range from January 1, 2010, to February 22, 2016, and the minimum time gap is one day. All intermediate events from February 22, 2012 to January 1, 2013 are absent, so there are 1931 days in total, and the training/validation/test split is 0.8/0.2/0.2. All experiments involving module training and inference are implemented on a local server with $4\times 48$ GB NVIDIA L40S GPUs. For detailed experimental set up, such as hyperparameters and instruction fine-tuning results for RoBERTa\textsubscript{BASE} encoder and FLAN-T5\textsubscript{BASE} generator, we discuss them in the appendix. 

\subsection{\model\textsubscript{OP1}: Ranking Results}

For ranking OP, we use Hits at $k$ as the evaluation metrics, which count the top$-k$ ranking positions of the correct objects among all candidates, where $k=1,3,10$. We consider comprehensive structural methods, including decoder-only ConvTransE~\cite{shang2019end}, GNN-based CompGCN~\cite{vashishth2019composition}, R-GCN~\cite{schlichtkrull2018modeling}, and RE-GCN~\cite{li2021temporal}, as well as hypergraph-based SeCoGD~\cite{ma2023context} involving pre-trained LDA context clustering~\cite{blei2003latent}, where the cluster number is set to 5. Although the scoring functions for entity ranking vary a lot, such as TransE~\cite{bordes2013translating}, TransR~\cite{lin2015learning}, and DistMult~\cite{yang2014embedding}, we uniformly apply ConvTransE, which is the SOTA option, towards all the aforementioned structural approaches, including \model\textsubscript{OP1}. 

Based on Table~\ref{tab:opranking}, we can observe that \model\textsubscript{OP1} achieves better prediction accuracy than other competitors among all three ranking levels. Additionally, by tracing back to a longer sequence, the overall scores will slightly increase, possibly due to abundant historical information being available. However, $l_1$, as a hyperparameter, cannot grow unlimitedly as the memory occupation to store more historical graphs and the training time to complete one epoch will correspondingly increase. 

\subsection{\model\textsubscript{OP2}: Generative Results}

For generative OP, we use ROUGE scores $(1/2/\text{L})$ as the evaluation metrics, which measures the overlap between generated and true objects at different levels including unigrams, bigrams, and longest common subsequence~\cite{lin2004rouge}. We leverage multiple generative LLMs, ranging from open-source Mistral-7B-Instruct \cite{mistral-ai} and Llama-3.1-8B-Instruct \cite{meta-llama31}, to closed-source GPT-3.5-Turbo-Instruct \cite{openai-gpt-35} and GPT-4o mini \cite{openai-gpt-4o}. Apart from generative models, we also incorporate competitive structural methods such as SeCoGD \cite{ma2023context} and \model\textsubscript{OP1}, by reading out the object entity which has the highest probability among all candidates. 

Based on Table~\ref{tab:opqa}, we can observe that \model\textsubscript{OP2} yields much better ROUGE scores even than \model\textsubscript{OP1}. Additionally, \model\textsubscript{OP2} with \textit{zero-shot prompt} achieves competitively similar scores as \textit{few-shot prompt}, demonstrating good in-domain generalizability of FLAN-T5\textsubscript{BASE} after QA fine-tuning. However, \model\textsubscript{OP2} with {no-text prompt} has significantly degraded scores across all three datasets, indicating that a textual summary is essential to enhance the contextualized understanding of a Seq2Seq LLM.

The fine-tuned FLAN-T5\textsubscript{BASE} is a great event interpolator. We also test direct prompting generative LLMs without any local training in Table~\ref{tab:opqa}. We believe that, under limited resources, it is difficult to obtain satisfying prediction accuracy simply through in-context learning without more advanced prompting strategies, as the results from both open-source and closed-source LLMs are similarly less competitive even than the structural \model\textsubscript{OP1}. Since most sociopolitical entities are countries or governmental institutions made of less than five words, the generative setting only needs to output a few words for every prediction, and therefore takes more advantage over the ranking formulation, which simultaneously evaluates hundreds of candidates. \textbf{To summarize OP}, if the entities to be predicted are short like ``Business (France)" and ``Businessperson (United States)", then the generative setting is a good choice. If the ground-truth targets involve long phrases, such as ``Cabinet / Council of Ministers / Advisors (Spain)", and there are dozens or even hundreds of long entities, then the ranking formulation is more suitable. 

\subsection{MEF: Multi-label Binary Classification Results}

For MEF, we utilize precision, recall, and F1 score to evaluate this multi-label binary classification task, where ``multi-label" indicates all unique relations $R$, and ``binary classification" represents either occurrence (1) or non-occurrence (0) of one specific relation $r_i\in R$, where $i=1,2,...,\lvert R\rvert$. 
Following the experimental setup in Glean~\cite{deng2020dynamic}, we consider comprehensive baselines including neural network-based MLARAM~\cite{benites2015haram} and DNN~\cite{bengio2009learning}, nearest neighbor-based BR-KNN~\cite{spyromitros2008empirical} and ML-KNN~\cite{zhang2007ml}, along with GNN and RNN-based DynamicGCN~\cite{deng2019learning}, TemporalGCN~\cite{zhao2019t}, and RENET~\cite{jin2019recurrent}. Compared with these SOTA methods, our approach \model\textsubscript{MEF} neither involve GNNs nor RNNs, therefore much design effort on updating entity and relation embeddings can be saved. 

Based on Table~\ref{table_MEF}, \model\textsubscript{MEF} achieves the best F1 score and recall all three datasets, and maintains a competitive precision among other baselines. We also conduct an ablation study by removing the self-attention (SA) layer \cite{vaswani2017attention}, denoted as \model\textsubscript{MEF\textbackslash SA}, which exhibits explicitly degraded forecasting accuracy. An intuitive take-away is that historical knowledge contains complicated relations among entities such as higher-order neighborhoods or hypergraph structure~\cite{ma2022learning}. \textbf{To summarize MEF}, we can observe that \model\textsubscript{MEF} achieves limited improvement in F1 score and precision, while the recall is significantly increased. This indicates our approach is good at imitating an observed trend during training, while the actual capability of forecasting the future is still constrained. 

\subsection{Computational Cost Comparison}

We further compare GPU memory occupation, training/inference time, and money spending among selected SOTA methods, using the Indian dataset. To report the training time for both \model\textsubscript{OP1} and \model\textsubscript{OP2}, we include the fine-tuning time for LLMs. As shown in Table~\ref{fig_object_prediction_generative} and Table~\ref{table_compare_cost}, \model\textsubscript{OP2} has the best event interpolation accuracy, but also takes the most locally significant training efforts. Meanwhile, GPT-4o mini achieves slightly better ROUGE scores with much cheaper cost than GPT-3.5 Turbo. For MEF, with the introduction of textual embeddings and self-attention, \model\textsubscript{MEF} is more computationally expensive than Glean, albeit an enhanced accuracy is achieved. 

\section{Discussion and Conclusion}

In this paper, we propose LEAP to transform event prediction as language understanding and reasoning tasks. We investigated LLM-based frameworks for both interpolated tasks such as object prediction and extrapolated tasks such as multi-event forecasting. We apply quintuple-level prompt engineering to address the challenge of limited input context of open-source LLMs, and fine-tune RoBERTa\textsubscript{BASE} as an encoder and FLAN-T5\textsubscript{BASE} as a generator to achieve state-of-the-art event prediction accuracy on multiple real-world datasets. With fine-tuned open-source LLMs, we achieve better prediction accuracy than commercialized generative LLMs on sociopolitical events. Furthermore, we discuss the limitations of our approaches and explore several new ideas as potential future work. 


The first limitation is that our OP and MEF approaches aim to simplify the design of existing event prediction frameworks~\cite{ma2023context,deng2020dynamic} by introducing either encoding or generative LLMs, with enhanced prediction accuracy and affordable development being our major focus. As time evolves, the capability of pre-trained LLMs to reason the development of sociopolitical events can be further explored by integrating more advanced prompting strategies such as Chain-of-Thought~\cite{wei2022chain} and ReAct~\cite{yao2022react}. 

The second limitation is that although multiple countries are involved, we only utilize ICEWS datasets \cite{DVN/28075_2015} for experiments, while other popular event datasets, such as GDELT \cite{leetaru2013gdelt} and YAGO \cite{pellissier2020yago}, are not included, as our approaches are based on well-formulated event quintuples instead of quadruples which belong to the typical definition of temporal events \cite{cai2022temporal}. In other words, extending a quadruple to a quintuple by introducing a brief text summary can be regarded as a retrieval augmentation strategy \cite{lewis2020retrieval}, which also demands more careful study. 

One promising future direction is to evaluate the capability of LLMs in reasoning and understanding human event development. This involves meticulous design of experiments incorporating human feedback. Another important direction is to enhance LLMs' temporal and spatiotemporal understanding of the world by integrating retrieval-augmented generation and knowledge graphs.

\section{Appendix: Experimental Setup}

In this section, we describe detailed hyperparameters for all sub-tasks of \model, and provide additional experimental results on LLM fine-tuning. Our experiments are implemented on a Ubuntu 22.04 server with 512GB RAM ($8\times 64\text{GB DDR5}$) and $4\times 48\text{GB}$ NVIDIA L40S GPUs. 

\subsection{\model\textsubscript{OP1}: Fine-tuning RoBERTa\textsubscript{BASE}}

We fine-tune RoBERTa\textsubscript{BASE} with masked language modeling (MLM) loss by concatenating all text summaries in the training split~\cite{liu2019roberta}. The learning rate is $2\times 10^{-5}$, the number of fine-tuning epochs is 40, the weight decay is $1\times 10^{-2}$, the block size for chunking the textual corpus is 512 tokens, and the masked language modeling probability is 0.15. We use perplexity to evaluate the MLM fine-tuning performance, as perplexity indicates how well a language model has learned the distribution of the text on which it has been trained~\cite{jelinek1977perplexity}. Based on Table~\ref{appendix_table_op1}, we observe an explicit decrease in test perplexity across all three datasets, which indicates that the fine-tuned RoBERTa\textsubscript{BASE} has enriched domain-specific knowledge. 

\begin{table*}[t]
\centering
\caption{Test ROUGE scores for FLAN-T5\textsubscript{BASE}, \textit{few-shot prompt}, $l_2=5$ quintuples.}
\label{appendix_table_op2}
\begin{tabular}{lccccccccccc}
\toprule
\textbf{Dataset} & \multicolumn{3}{c}{\textbf{Afghanistan}} &&  \multicolumn{3}{c}{\textbf{India}} &&  \multicolumn{3}{c}{\textbf{Russia}} \\
\midrule
ROUGE$-$  & 1      & 2      & L     &  & 1      & 2      & L &      & 1      & 2      & L \\
\midrule
Before fine-tuning & 0.2082 & 0.0077 & 0.2079 & &  
                     0.0531 & 0.0194 & 0.0531 & &  
                     0.1370 & 0.0151 & 0.1369 \\
After fine-tuning & \textbf{0.8638} & \textbf{0.5656} & \textbf{0.8638} & & 
                    \textbf{0.8594} & \textbf{0.6962} & \textbf{0.8594} & & 
                    \textbf{0.8415} & \textbf{0.4544} & \textbf{0.8414} \\
\bottomrule
\end{tabular}
\end{table*}

\begin{table*}[t]
\centering
\caption{Test MEF results over 5 runs, $\text{mean}\pm \text{standard deviation}$, $l_3=7$ days.}
\label{appendix_table_mef}
\begin{tabular}{lcccccccc}
\toprule
\textbf{Dataset} & \multicolumn{2}{c}{\textbf{Afghanistan}} & &
                   \multicolumn{2}{c}{\textbf{India}} & &
                   \multicolumn{2}{c}{\textbf{Russia}} \\
\midrule
 & F1 score & Recall & &
   F1 score & Recall & &
   F1 score & Recall \\
\midrule
Glean & 
$\text{0.6259}_{\pm \text{0.0013}}$ & 
$\text{0.8341}_{\pm \text{0.0059}}$ & &
$\text{0.6623}_{\pm \text{0.0064}}$ & 
$\text{0.7646}_{\pm \text{0.0107}}$ & &
$\text{0.5972}_{\pm \text{0.0082}}$ & 
$\text{0.7606}_{\pm \text{0.0258}}$ \\
\model\textsubscript{MEF} & 
$\textbf{0.6367}_{\pm \textbf{0.0003}}$ & 
$\textbf{0.8894}_{\pm \textbf{0.0013}}$ & &
$\textbf{0.7081}_{\pm \textbf{0.0019}}$ & 
$\textbf{0.8684}_{\pm \textbf{0.0048}}$ & &
$\textbf{0.6279}_{\pm \textbf{0.0002}}$ & 
$\textbf{0.8673}_{\pm \textbf{0.0014}}$ \\
\midrule
Gain\textsubscript{Absolute} & 
0.0108 & 0.0553 & & 
0.0458 & 0.1038 & & 
0.0307 & 0.1067 \\
Gain\textsubscript{Relative} & 
1.725\% & 6.630\% & & 
6.915\% & 13.576\% & & 
5.141\% & 14.028\% \\
\bottomrule
\end{tabular}
\end{table*}

\begin{table}[t]
\centering
\caption{Test perplexities for RoBERTa\textsubscript{BASE}.}
\label{appendix_table_op1}
\begin{tabular}{lccc}
\toprule
\textbf{Dataset} & \textbf{Afghanistan}
                 & \textbf{India}
                 & \textbf{Russia} \\
\midrule
Before fine-tuning & 3.53 & 3.65 & 2.85 \\
After fine-tuning & \textbf{2.03} & \textbf{2.32} & \textbf{2.01} \\
\bottomrule
\end{tabular}
\end{table}

\begin{table}[t]
\centering
\caption{Wilcoxon signed-rank tests for MEF results.}
\label{appendix_table_mef_stat}
\begin{tabular}{lccc}
\toprule
\textbf{Metric} & \textbf{F1 score}
                 & \textbf{Recall}
                 & \textbf{Precision} \\
\midrule
$\text{p}-$value\textsubscript{one-sided} & $3\times 10^{-5}$ & $3\times 10^{-5}$ & 0.1947 \\
$\text{p}-$value\textsubscript{two-sided} & $6\times 10^{-5}$ & $6\times 10^{-5}$ & 0.3894 \\
\bottomrule
\end{tabular}
\end{table}

\subsection{\model\textsubscript{OP1}: Training Structural Modules}

To jointly train the structural modules in \model\textsubscript{OP1} including R-GCN~\cite{schlichtkrull2018modeling}, GRU~\cite{cho2014learning}, and ConvTransE~\cite{shang2019end}, we leverage cross-entropy loss and set the number of training epochs to 40 with a patience equal to 5 epochs for early stopping. R-GCN has 2 layers, and the dropout rate is 0.2. The learning rate is $1\times 10^{-3}$, the weight decay is $1\times 10^{-6}$, the batch size is 1 (day), and the normalization threshold for gradient clipping is 1.0, along with the Adam optimizer~\cite{kingma2014adam}. The feature dimension of entity and relation embeddings is 200, while the dimension for text summary embeddings from the fine-tuned RoBERTa\textsubscript{BASE} is 50265. 

\subsection{\model\textsubscript{OP2}: Fine-tuning FLAN-T5\textsubscript{BASE}}

We fine-tune FLAN-T5\textsubscript{BASE} with cross-entropy loss, which is iteratively calculated by comparing the next generated token with the ground-truth token~\cite{chung2024scaling}. The learning rate is $3\times 10^{-4}$, the number of fine-tuning epochs is 5, the weight decay is $1\times 10^{-2}$, and the batch size is 2 (prompts). We use ROUGE scores~\cite{lin2004rouge} to evaluate the fine-tuning performance, as they can be computed by comparing the generated objects with the ground-truth objects, so that the object prediction quality can be simultaneously evaluated. Based on Table~\ref{appendix_table_op2}, we can observe that the ROUGE scores are significantly increased after fine-tuning, which demonstrates that the fine-tuned FLAN-T5\textsubscript{BASE} is a more capable generative interpolator. 

\subsection{Training and Evaluating \model\textsubscript{MEF}}

To jointly train the self-attention~\cite{vaswani2017attention} and feature projection layers in \model\textsubscript{MEF}, we apply cross-entropy loss and set the number of training epochs to 40 with a patience equal to 5 epochs for early stopping. The learning rate is $5\times 10^{-5}$, the weight decay is $1\times 10^{-2}$, the batch size is 2 (days), and the normalization threshold for gradient clipping is 1.0, along with the Adam optimizer~\cite{kingma2014adam}. The feature dimension for quintuple embeddings from the pre-trained RoBERTa\textsubscript{LARGE} (without fine-tuning) is 50265, and the feature dimension after quintuple embedding aggregation and projection is 1024. 

In the main paper, all event prediction experiments are implemented over 1 run. In this subsection, we implement Glean~\cite{deng2020dynamic} and \model\textsubscript{MEF} over 5 runs and report the mean and standard deviation values of F1 score and recall. We also compute the absolute and relative gains with respect to the mean values. Based on Table~\ref{appendix_table_mef}, we can observe that \model\textsubscript{MEF} achieves better forecasting accuracy and stability than Glean, with larger means and smaller standard deviations. 

In addition, we conduct both one-sided and two-sided Wilcoxon signed-rank tests~\cite{wilcoxon1992individual} for MEF results over 5 runs and across 3 datasets, which leads to 15 paired samples between \model\textsubscript{MEF} and Glean for each test. As shown in Table~\ref{appendix_table_mef_stat}, we can conclude that \model\textsubscript{MEF} explicitly outperforms Glean in terms of F1 score and recall, as their one-sided and two-sided $\text{p}-$values are much smaller than 0.05. However, the multi-event forecasting performance difference, when evaluated by precision, is not significant as neither $\text{p}-$values are less than 0.05. This indicates that \model\textsubscript{MEF} still exhibits a limited capability of precisely forecasting events in the future. 


\begin{thebibliography}{59}
\providecommand{\natexlab}[1]{#1}

\bibitem[{Anthropic(2024)}]{anthropic-claude-3}
Anthropic. 2024.
\newblock Introducing the next generation of Claude.
\newblock \url{https://www.anthropic.com/news/claude-3-family}.
\newblock Accessed: 2024-08-05.

\bibitem[{Bengio et~al.(2009)}]{bengio2009learning}
Bengio, Y.; et~al. 2009.
\newblock Learning deep architectures for AI.
\newblock \emph{Foundations and trends{\textregistered} in Machine Learning}, 2(1): 1--127.

\bibitem[{Benites and Sapozhnikova(2015)}]{benites2015haram}
Benites, F.; and Sapozhnikova, E. 2015.
\newblock Haram: a hierarchical aram neural network for large-scale text classification.
\newblock In \emph{2015 IEEE international conference on data mining workshop (ICDMW)}, 847--854. IEEE.

\bibitem[{Blei, Ng, and Jordan(2003)}]{blei2003latent}
Blei, D.~M.; Ng, A.~Y.; and Jordan, M.~I. 2003.
\newblock Latent dirichlet allocation.
\newblock \emph{Journal of machine Learning research}, 3(Jan): 993--1022.

\bibitem[{Bordes et~al.(2013)Bordes, Usunier, Garcia-Duran, Weston, and Yakhnenko}]{bordes2013translating}
Bordes, A.; Usunier, N.; Garcia-Duran, A.; Weston, J.; and Yakhnenko, O. 2013.
\newblock Translating embeddings for modeling multi-relational data.
\newblock \emph{Advances in neural information processing systems}, 26.

\bibitem[{Boschee et~al.(2015)Boschee, Lautenschlager, O'Brien, Shellman, Starz, and Ward}]{DVN/28075_2015}
Boschee, E.; Lautenschlager, J.; O'Brien, S.; Shellman, S.; Starz, J.; and Ward, M. 2015.
\newblock {ICEWS Coded Event Data}.

\bibitem[{Cai et~al.(2022)Cai, Xiang, Gao, Zhang, Li, and Li}]{cai2022temporal}
Cai, B.; Xiang, Y.; Gao, L.; Zhang, H.; Li, Y.; and Li, J. 2022.
\newblock Temporal knowledge graph completion: A survey.
\newblock \emph{arXiv preprint arXiv:2201.08236}.

\bibitem[{Cai et~al.(2024)Cai, Mao, Zhou, Long, Wu, and Lan}]{cai2024survey}
Cai, L.; Mao, X.; Zhou, Y.; Long, Z.; Wu, C.; and Lan, M. 2024.
\newblock A Survey on Temporal Knowledge Graph: Representation Learning and Applications.
\newblock \emph{arXiv preprint arXiv:2403.04782}.

\bibitem[{Cho et~al.(2014)Cho, Van~Merri{\"e}nboer, Gulcehre, Bahdanau, Bougares, Schwenk, and Bengio}]{cho2014learning}
Cho, K.; Van~Merri{\"e}nboer, B.; Gulcehre, C.; Bahdanau, D.; Bougares, F.; Schwenk, H.; and Bengio, Y. 2014.
\newblock Learning phrase representations using RNN encoder-decoder for statistical machine translation.
\newblock \emph{arXiv preprint arXiv:1406.1078}.

\bibitem[{Chung et~al.(2024)Chung, Hou, Longpre, Zoph, Tay, Fedus, Li, Wang, Dehghani, Brahma et~al.}]{chung2024scaling}
Chung, H.~W.; Hou, L.; Longpre, S.; Zoph, B.; Tay, Y.; Fedus, W.; Li, Y.; Wang, X.; Dehghani, M.; Brahma, S.; et~al. 2024.
\newblock Scaling instruction-finetuned language models.
\newblock \emph{Journal of Machine Learning Research}, 25(70): 1--53.

\bibitem[{Church and Hanks(1990)}]{church1990word}
Church, K.; and Hanks, P. 1990.
\newblock Word association norms, mutual information, and lexicography.
\newblock \emph{Computational linguistics}, 16(1): 22--29.

\bibitem[{Deng, Rangwala, and Ning(2019)}]{deng2019learning}
Deng, S.; Rangwala, H.; and Ning, Y. 2019.
\newblock Learning dynamic context graphs for predicting social events.
\newblock In \emph{Proceedings of the 25th ACM SIGKDD International Conference on Knowledge Discovery \& Data Mining}, 1007--1016.

\bibitem[{Deng, Rangwala, and Ning(2020)}]{deng2020dynamic}
Deng, S.; Rangwala, H.; and Ning, Y. 2020.
\newblock Dynamic knowledge graph based multi-event forecasting.
\newblock In \emph{Proceedings of the 26th ACM SIGKDD International Conference on Knowledge Discovery \& Data Mining}, 1585--1595.

\bibitem[{Dong et~al.(2022)Dong, Li, Dai, Zheng, Wu, Chang, Sun, Xu, and Sui}]{dong2022survey}
Dong, Q.; Li, L.; Dai, D.; Zheng, C.; Wu, Z.; Chang, B.; Sun, X.; Xu, J.; and Sui, Z. 2022.
\newblock A survey on in-context learning.
\newblock \emph{arXiv preprint arXiv:2301.00234}.

\bibitem[{Jelinek et~al.(1977)Jelinek, Mercer, Bahl, and Baker}]{jelinek1977perplexity}
Jelinek, F.; Mercer, R.~L.; Bahl, L.~R.; and Baker, J.~K. 1977.
\newblock Perplexity—a measure of the difficulty of speech recognition tasks.
\newblock \emph{The Journal of the Acoustical Society of America}, 62(S1): S63--S63.

\bibitem[{Jin et~al.(2019)Jin, Qu, Jin, and Ren}]{jin2019recurrent}
Jin, W.; Qu, M.; Jin, X.; and Ren, X. 2019.
\newblock Recurrent event network: Autoregressive structure inference over temporal knowledge graphs.
\newblock \emph{arXiv preprint arXiv:1904.05530}.

\bibitem[{Kanakarajan and Sankarasubbu(2023)}]{kanakarajan2023saama}
Kanakarajan, K.~R.; and Sankarasubbu, M. 2023.
\newblock Saama AI Research at SemEval-2023 Task 7: Exploring the Capabilities of Flan-T5 for Multi-evidence Natural Language Inference in Clinical Trial Data.
\newblock In \emph{Proceedings of the 17th International Workshop on Semantic Evaluation (SemEval-2023)}, 995--1003.

\bibitem[{Kingma(2014)}]{kingma2014adam}
Kingma, D.~P. 2014.
\newblock Adam: A method for stochastic optimization.
\newblock \emph{arXiv preprint arXiv:1412.6980}.

\bibitem[{Lee et~al.(2023)Lee, Ahrabian, Jin, Morstatter, and Pujara}]{lee2023temporal}
Lee, D.-H.; Ahrabian, K.; Jin, W.; Morstatter, F.; and Pujara, J. 2023.
\newblock Temporal knowledge graph forecasting without knowledge using in-context learning.
\newblock \emph{arXiv preprint arXiv:2305.10613}.

\bibitem[{Leetaru and Schrodt(2013)}]{leetaru2013gdelt}
Leetaru, K.; and Schrodt, P.~A. 2013.
\newblock Gdelt: Global data on events, location, and tone, 1979--2012.
\newblock In \emph{ISA annual convention}, volume~2, 1--49. Citeseer.

\bibitem[{Lewis et~al.(2019)Lewis, Liu, Goyal, Ghazvininejad, Mohamed, Levy, Stoyanov, and Zettlemoyer}]{lewis2019bart}
Lewis, M.; Liu, Y.; Goyal, N.; Ghazvininejad, M.; Mohamed, A.; Levy, O.; Stoyanov, V.; and Zettlemoyer, L. 2019.
\newblock Bart: Denoising sequence-to-sequence pre-training for natural language generation, translation, and comprehension.
\newblock \emph{arXiv preprint arXiv:1910.13461}.

\bibitem[{Lewis et~al.(2020)Lewis, Perez, Piktus, Petroni, Karpukhin, Goyal, K{\"u}ttler, Lewis, Yih, Rockt{\"a}schel et~al.}]{lewis2020retrieval}
Lewis, P.; Perez, E.; Piktus, A.; Petroni, F.; Karpukhin, V.; Goyal, N.; K{\"u}ttler, H.; Lewis, M.; Yih, W.-t.; Rockt{\"a}schel, T.; et~al. 2020.
\newblock Retrieval-augmented generation for knowledge-intensive nlp tasks.
\newblock \emph{Advances in Neural Information Processing Systems}, 33: 9459--9474.

\bibitem[{Li et~al.(2021)Li, Jin, Li, Guan, Guo, Shen, Wang, and Cheng}]{li2021temporal}
Li, Z.; Jin, X.; Li, W.; Guan, S.; Guo, J.; Shen, H.; Wang, Y.; and Cheng, X. 2021.
\newblock Temporal knowledge graph reasoning based on evolutional representation learning.
\newblock In \emph{Proceedings of the 44th international ACM SIGIR conference on research and development in information retrieval}, 408--417.

\bibitem[{Liao et~al.(2023)Liao, Jia, Ma, and Tresp}]{liao2023gentkg}
Liao, R.; Jia, X.; Ma, Y.; and Tresp, V. 2023.
\newblock GenTKG: Generative Forecasting on Temporal Knowledge Graph.
\newblock \emph{arXiv preprint arXiv:2310.07793}.

\bibitem[{Lin(2004)}]{lin2004rouge}
Lin, C.-Y. 2004.
\newblock Rouge: A package for automatic evaluation of summaries.
\newblock In \emph{Text summarization branches out}, 74--81.

\bibitem[{Lin et~al.(2015)Lin, Liu, Sun, Liu, and Zhu}]{lin2015learning}
Lin, Y.; Liu, Z.; Sun, M.; Liu, Y.; and Zhu, X. 2015.
\newblock Learning entity and relation embeddings for knowledge graph completion.
\newblock In \emph{Proceedings of the AAAI conference on artificial intelligence}, volume~29.

\bibitem[{Liu et~al.(2022)Liu, Ma, Hildebrandt, Joblin, and Tresp}]{liu2022tlogic}
Liu, Y.; Ma, Y.; Hildebrandt, M.; Joblin, M.; and Tresp, V. 2022.
\newblock Tlogic: Temporal logical rules for explainable link forecasting on temporal knowledge graphs.
\newblock In \emph{Proceedings of the AAAI conference on artificial intelligence}, volume~36, 4120--4127.

\bibitem[{Liu et~al.(2019)Liu, Ott, Goyal, Du, Joshi, Chen, Levy, Lewis, Zettlemoyer, and Stoyanov}]{liu2019roberta}
Liu, Y.; Ott, M.; Goyal, N.; Du, J.; Joshi, M.; Chen, D.; Levy, O.; Lewis, M.; Zettlemoyer, L.; and Stoyanov, V. 2019.
\newblock Roberta: A robustly optimized bert pretraining approach.
\newblock \emph{arXiv preprint arXiv:1907.11692}.

\bibitem[{Ma et~al.(2022)Ma, Wan, Yang, Li, Hecht, and Teevan}]{ma2022learning}
Ma, J.; Wan, M.; Yang, L.; Li, J.; Hecht, B.; and Teevan, J. 2022.
\newblock Learning causal effects on hypergraphs.
\newblock In \emph{Proceedings of the 28th ACM SIGKDD Conference on Knowledge Discovery and Data Mining}, 1202--1212.

\bibitem[{Ma et~al.(2023)Ma, Ye, Wu, Wang, Cao, and Chua}]{ma2023context}
Ma, Y.; Ye, C.; Wu, Z.; Wang, X.; Cao, Y.; and Chua, T.-S. 2023.
\newblock Context-aware event forecasting via graph disentanglement.
\newblock In \emph{Proceedings of the 29th ACM SIGKDD Conference on Knowledge Discovery and Data Mining}, 1643--1652.

\bibitem[{Meta(2024)}]{meta-llama31}
Meta. 2024.
\newblock Introducing Llama 3.1: Our most capable models to date.
\newblock \url{https://ai.meta.com/blog/meta-llama-3-1/}.
\newblock Accessed: 2024-08-05.

\bibitem[{MistralAI(2023)}]{mistral-ai}
MistralAI. 2023.
\newblock Mistral 7B: The best 7B model to date, Apache 2.0.
\newblock \url{https://mistral.ai/news/announcing-mistral-7b/}.
\newblock Accessed: 2024-08-05.

\bibitem[{Onaolapo et~al.(2022)Onaolapo, Carpanen, Dorrell, and Ojo}]{onaolapo2022event}
Onaolapo, A.~K.; Carpanen, R.~P.; Dorrell, D.~G.; and Ojo, E.~E. 2022.
\newblock Event-driven power outage prediction using collaborative neural networks.
\newblock \emph{IEEE Transactions on Industrial Informatics}, 19(3): 3079--3087.

\bibitem[{OpenAI(2023)}]{openai-gpt-35}
OpenAI. 2023.
\newblock GPT-3.5 Turbo.
\newblock \url{https://platform.openai.com/docs/models/gpt-3-5-turbo}.
\newblock Accessed: 2024-08-05.

\bibitem[{OpenAI(2024{\natexlab{a}})}]{openai-gpt-4o}
OpenAI. 2024{\natexlab{a}}.
\newblock Hello GPT-4o.
\newblock \url{https://openai.com/index/hello-gpt-4o/}.
\newblock Accessed: 2024-08-05.

\bibitem[{OpenAI(2024{\natexlab{b}})}]{openai-pricing}
OpenAI. 2024{\natexlab{b}}.
\newblock Pricing.
\newblock \url{https://openai.com/api/pricing/}.
\newblock Accessed: 2024-08-05.

\bibitem[{Ouyang et~al.(2022)Ouyang, Wu, Jiang, Almeida, Wainwright, Mishkin, Zhang, Agarwal, Slama, Ray et~al.}]{ouyang2022training}
Ouyang, L.; Wu, J.; Jiang, X.; Almeida, D.; Wainwright, C.; Mishkin, P.; Zhang, C.; Agarwal, S.; Slama, K.; Ray, A.; et~al. 2022.
\newblock Training language models to follow instructions with human feedback.
\newblock \emph{Advances in neural information processing systems}, 35: 27730--27744.

\bibitem[{Pan et~al.(2023)Pan, Nayyeri, Li, and Staab}]{pan2023temporal}
Pan, J.; Nayyeri, M.; Li, Y.; and Staab, S. 2023.
\newblock Do Temporal Knowledge Graph Embedding Models Learn or Memorize Shortcuts?
\newblock In \emph{Temporal Graph Learning Workshop@ NeurIPS 2023}.

\bibitem[{Pan et~al.(2024)Pan, Luo, Wang, Chen, Wang, and Wu}]{pan2024unifying}
Pan, S.; Luo, L.; Wang, Y.; Chen, C.; Wang, J.; and Wu, X. 2024.
\newblock Unifying large language models and knowledge graphs: A roadmap.
\newblock \emph{IEEE Transactions on Knowledge and Data Engineering}.

\bibitem[{Pellissier~Tanon, Weikum, and Suchanek(2020)}]{pellissier2020yago}
Pellissier~Tanon, T.; Weikum, G.; and Suchanek, F. 2020.
\newblock Yago 4: A reason-able knowledge base.
\newblock In \emph{The Semantic Web: 17th International Conference, ESWC 2020, Heraklion, Crete, Greece, May 31--June 4, 2020, Proceedings 17}, 583--596. Springer.

\bibitem[{Schlichtkrull et~al.(2018)Schlichtkrull, Kipf, Bloem, Van Den~Berg, Titov, and Welling}]{schlichtkrull2018modeling}
Schlichtkrull, M.; Kipf, T.~N.; Bloem, P.; Van Den~Berg, R.; Titov, I.; and Welling, M. 2018.
\newblock Modeling relational data with graph convolutional networks.
\newblock In \emph{The semantic web: 15th international conference, ESWC 2018, Heraklion, Crete, Greece, June 3--7, 2018, proceedings 15}, 593--607. Springer.

\bibitem[{Shang et~al.(2019)Shang, Tang, Huang, Bi, He, and Zhou}]{shang2019end}
Shang, C.; Tang, Y.; Huang, J.; Bi, J.; He, X.; and Zhou, B. 2019.
\newblock End-to-end structure-aware convolutional networks for knowledge base completion.
\newblock In \emph{Proceedings of the AAAI conference on artificial intelligence}, volume~33, 3060--3067.

\bibitem[{Shang and Huang(2024)}]{shang2024survey}
Shang, W.; and Huang, X. 2024.
\newblock A Survey of Large Language Models on Generative Graph Analytics: Query, Learning, and Applications.
\newblock \emph{arXiv preprint arXiv:2404.14809}.

\bibitem[{Shi et~al.(2024)Shi, Xue, Wang, Zhou, Zhang, Zhou, Tan, and Mei}]{shi2024language}
Shi, X.; Xue, S.; Wang, K.; Zhou, F.; Zhang, J.; Zhou, J.; Tan, C.; and Mei, H. 2024.
\newblock Language models can improve event prediction by few-shot abductive reasoning.
\newblock \emph{Advances in Neural Information Processing Systems}, 36.

\bibitem[{Spyromitros, Tsoumakas, and Vlahavas(2008)}]{spyromitros2008empirical}
Spyromitros, E.; Tsoumakas, G.; and Vlahavas, I. 2008.
\newblock An empirical study of lazy multilabel classification algorithms.
\newblock In \emph{Artificial Intelligence: Theories, Models and Applications: 5th Hellenic Conference on AI, SETN 2008, Syros, Greece, October 2-4, 2008. Proceedings 5}, 401--406. Springer.

\bibitem[{Touvron et~al.(2023{\natexlab{a}})Touvron, Lavril, Izacard, Martinet, Lachaux, Lacroix, Rozi{\`e}re, Goyal, Hambro, Azhar et~al.}]{touvron2023llama}
Touvron, H.; Lavril, T.; Izacard, G.; Martinet, X.; Lachaux, M.-A.; Lacroix, T.; Rozi{\`e}re, B.; Goyal, N.; Hambro, E.; Azhar, F.; et~al. 2023{\natexlab{a}}.
\newblock Llama: Open and efficient foundation language models.
\newblock \emph{arXiv preprint arXiv:2302.13971}.

\bibitem[{Touvron et~al.(2023{\natexlab{b}})Touvron, Martin, Stone, Albert, Almahairi, Babaei, Bashlykov, Batra, Bhargava, Bhosale et~al.}]{touvron2023llama2}
Touvron, H.; Martin, L.; Stone, K.; Albert, P.; Almahairi, A.; Babaei, Y.; Bashlykov, N.; Batra, S.; Bhargava, P.; Bhosale, S.; et~al. 2023{\natexlab{b}}.
\newblock Llama 2: Open foundation and fine-tuned chat models.
\newblock \emph{arXiv preprint arXiv:2307.09288}.

\bibitem[{Vashishth et~al.(2019)Vashishth, Sanyal, Nitin, and Talukdar}]{vashishth2019composition}
Vashishth, S.; Sanyal, S.; Nitin, V.; and Talukdar, P. 2019.
\newblock Composition-based multi-relational graph convolutional networks.
\newblock \emph{arXiv preprint arXiv:1911.03082}.

\bibitem[{Vaswani et~al.(2017)Vaswani, Shazeer, Parmar, Uszkoreit, Jones, Gomez, Kaiser, and Polosukhin}]{vaswani2017attention}
Vaswani, A.; Shazeer, N.; Parmar, N.; Uszkoreit, J.; Jones, L.; Gomez, A.~N.; Kaiser, {\L}.; and Polosukhin, I. 2017.
\newblock Attention is all you need.
\newblock \emph{Advances in neural information processing systems}, 30.

\bibitem[{Wang et~al.(2023)Wang, Wang, Qiu, Pan, Xiong, Liu, Luo, Liu, Hu, Yin et~al.}]{wang2023survey}
Wang, J.; Wang, B.; Qiu, M.; Pan, S.; Xiong, B.; Liu, H.; Luo, L.; Liu, T.; Hu, Y.; Yin, B.; et~al. 2023.
\newblock A survey on temporal knowledge graph completion: Taxonomy, progress, and prospects.
\newblock \emph{arXiv preprint arXiv:2308.02457}.

\bibitem[{Wasserkrug(2009)}]{Wasserkrug2009}
Wasserkrug, S. 2009.
\newblock \emph{Event Prediction}, 1048--1052.
\newblock Boston, MA: Springer US.
\newblock ISBN 978-0-387-39940-9.

\bibitem[{Wei et~al.(2022)Wei, Wang, Schuurmans, Bosma, Xia, Chi, Le, Zhou et~al.}]{wei2022chain}
Wei, J.; Wang, X.; Schuurmans, D.; Bosma, M.; Xia, F.; Chi, E.; Le, Q.~V.; Zhou, D.; et~al. 2022.
\newblock Chain-of-thought prompting elicits reasoning in large language models.
\newblock \emph{Advances in neural information processing systems}, 35: 24824--24837.

\bibitem[{Wilcoxon(1992)}]{wilcoxon1992individual}
Wilcoxon, F. 1992.
\newblock Individual comparisons by ranking methods.
\newblock In \emph{Breakthroughs in statistics: Methodology and distribution}, 196--202. Springer.

\bibitem[{Yang et~al.(2014)Yang, Yih, He, Gao, and Deng}]{yang2014embedding}
Yang, B.; Yih, W.-t.; He, X.; Gao, J.; and Deng, L. 2014.
\newblock Embedding entities and relations for learning and inference in knowledge bases.
\newblock \emph{arXiv preprint arXiv:1412.6575}.

\bibitem[{Yao et~al.(2022)Yao, Zhao, Yu, Du, Shafran, Narasimhan, and Cao}]{yao2022react}
Yao, S.; Zhao, J.; Yu, D.; Du, N.; Shafran, I.; Narasimhan, K.; and Cao, Y. 2022.
\newblock React: Synergizing reasoning and acting in language models.
\newblock \emph{arXiv preprint arXiv:2210.03629}.

\bibitem[{Ye et~al.(2024)Ye, Hu, Deng, Huang, Ma, Zhu, and Wang}]{ye2024mirai}
Ye, C.; Hu, Z.; Deng, Y.; Huang, Z.; Ma, M.~D.; Zhu, Y.; and Wang, W. 2024.
\newblock MIRAI: Evaluating LLM Agents for Event Forecasting.
\newblock \emph{arXiv preprint arXiv:2407.01231}.

\bibitem[{Zhang and Zhou(2007)}]{zhang2007ml}
Zhang, M.-L.; and Zhou, Z.-H. 2007.
\newblock ML-KNN: A lazy learning approach to multi-label learning.
\newblock \emph{Pattern recognition}, 40(7): 2038--2048.

\bibitem[{Zhao(2021)}]{zhao2021event}
Zhao, L. 2021.
\newblock Event prediction in the big data era: A systematic survey.
\newblock \emph{ACM Computing Surveys (CSUR)}, 54(5): 1--37.

\bibitem[{Zhao et~al.(2019)Zhao, Song, Zhang, Liu, Wang, Lin, Deng, and Li}]{zhao2019t}
Zhao, L.; Song, Y.; Zhang, C.; Liu, Y.; Wang, P.; Lin, T.; Deng, M.; and Li, H. 2019.
\newblock T-gcn: A temporal graph convolutional network for traffic prediction.
\newblock \emph{IEEE transactions on intelligent transportation systems}, 21(9): 3848--3858.

\end{thebibliography}

\end{document}